\documentclass[letterpaper, 10 pt, conference]{ieeeconf}  % Comment this line out if you need a4paper
\IEEEoverridecommandlockouts                              % This command is only needed if
\usepackage{graphics}    % for pdf, bitmapped graphics files
\usepackage{times}       % assumes new font selection scheme installed
\usepackage{amsmath}     % assumes amsmath package installed
\usepackage{amssymb}     % assumes amsmath package installed
\usepackage{graphicx}
\usepackage{algorithm}
\usepackage[noend]{algpseudocode}
\usepackage{siunitx}
\usepackage{dsfont}
\usepackage{bm}
\usepackage{booktabs}
\usepackage{subfig}
\usepackage{graphicx}
\usepackage{hyperref}
\usepackage[font=small]{caption}

\def\figref#1{Fig.~\ref{#1}}
\def\tabref#1{Tab.~\ref{#1}}
\def\eqref#1{Eq.~(\ref{#1})}

\newcommand\etal{\emph{et al.}}

\title{\LARGE \bf Efficient Manipulation-Enhanced Semantic Mapping \\ With Uncertainty-Informed Action Selection}
\author{Nils Dengler \and Jesper Mücke \and Rohit Menon \and Maren Bennewitz% <-this % stops a space
\thanks{ All authors are with the University of Bonn, Germany. 
   M. Bennewitz, R. Menon, and N. Dengler are additionally with the Lamarr Institute for Machine Learning and Artificial Intelligence with the Center for Robotics, Bonn, Germany. 
   This work has partly been supported by the European Commission under grant agreement numbers 964854 (RePAIR) and by the BMBF within the Robotics Institute Germany, grant No. 16ME0999.
}
}
\begin{document}
\maketitle
\thispagestyle{empty} 
\pagestyle{empty}

%%%%%%%%%%%%%%%%%%%%%%%%%%%%%%%%%%%%%%%%%%%%%%%%%%%%%%%%%%%%%%%%%%%%%%%%%%%%%%%%
\begin{abstract} 
Service robots operating in cluttered human environments such as homes, offices, and schools cannot rely on predefined object arrangements and must continuously update their semantic and spatial estimates while dealing with possible frequent rearrangements. 
Efficient and accurate mapping under such conditions demands selecting informative viewpoints and targeted manipulations to reduce occlusions and uncertainty. 
In this work, we present a manipulation-enhanced semantic mapping framework for occlusion-heavy shelf scenes that integrates evidential metric-semantic mapping with reinforcement-learning-based next-best view planning and targeted action selection. 
Our method thereby exploits uncertainty estimates from Dirichlet and Beta distributions in the map prediction networks to guide both active sensor placement and object manipulation, focusing on areas with high uncertainty and selecting actions with high expected information gain. 
Furthermore, we introduce an uncertainty-informed push strategy that targets occlusion-critical objects and generates minimally invasive actions to reveal hidden regions by reducing overall uncertainty in the scene. 
The experimental evaluation shows that our framework enables to accurately map cluttered scenes, while substantially reducing object displacement and achieving a 95\%~reduction in planning time compared to the state-of-the-art, thereby realizing real-world applicability.
\end{abstract}

%%%%%%%%%%%%%%%%%%%%%%%%%%%%%%%%%%%%%%%%%%%%%%%%%%%%%%%%%%%%%%%%%%%%%%%%%%%%%%%%
\section{Introduction}
\label{sec:intro}
The successful deployment of general-purpose service robots in homes and offices relies on perceiving and manipulating diverse objects in cluttered and constrained spaces. Robots must move beyond passive perception (e.g., mapping with a static camera) and instead apply active perception~\cite{o2019computer} in combination with deliberate physical object interactions~\cite{jiang2024roboexp}. Interactive perception techniques can remove occlusions by revealing hidden regions~\cite{huang2022mechanical,dengler23iros}, which is crucial for exploring unknown environments~\cite{engelbracht2024spotlight}, building detailed object models~\cite{yao2025icra}, and locating occluded objects~\cite{miao2022safe}.
However, the involved manipulation actions come at a cost, as uninformed interactions can increase uncertainty in the target object’s pose and stability, as well as in the neighboring scene elements. Hence, a central challenge in interactive perception is uncertainty-aware action selection.

To address this challenge, predictive networks have been used to estimate the effects of pushes in tabletop clutter for object retrieval and grasp–push selection~\cite{tang2023selective,huang2021visual}, but they target perceptually simple setups and ignore uncertainty. Marques~\etal~\cite{marques25rss} showed that evidential networks improve \emph{Manipulation-Enhanced Mapping} (MEM) in confined shelves by jointly predicting post-action occupancy and semantic beliefs.
However, they do not leverage belief uncertainty for continuous action planning.
Therefore, we propose three improvements: (i) continuous-space next-best-view (NBV) planning for fine-grained sensor placement that better reveals occlusions; (ii) uncertainty-informed push sampling that selects candidates by expected impact on hidden regions, improving efficiency and safety, i.e., reducing collisions with other objects; and (iii) NBV exploration driven by semantic uncertainty rather than occupancy entropy alone, improving both occupancy and semantic beliefs.

\begin{figure}[t]
\centering
\includegraphics[width=0.95\linewidth,trim={0.cm 1.5cm 1cm .5cm}, clip]{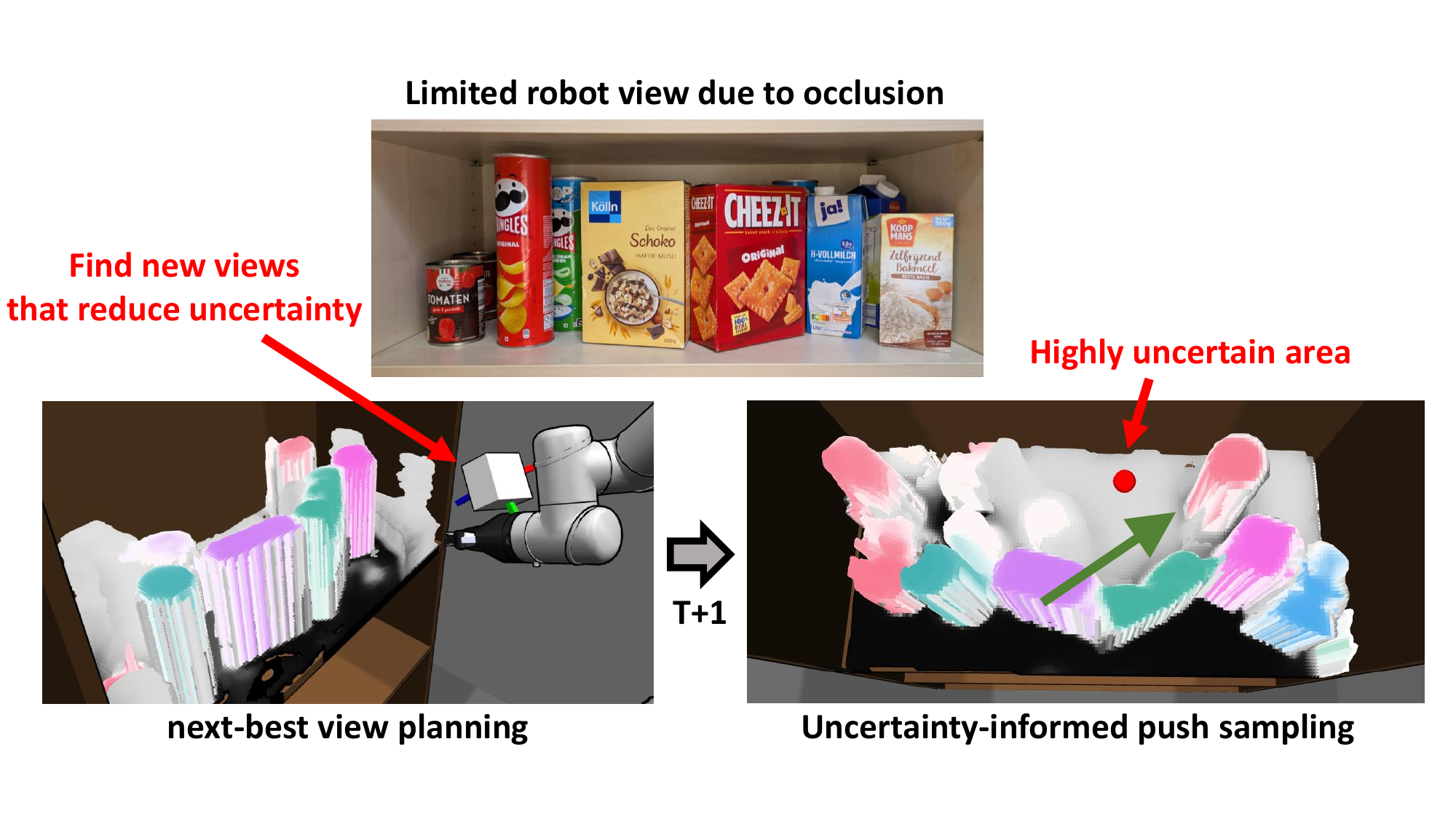}
%needed map t-1 side by side with real environment
% maybe use fov polygon in map to show the problem
\caption{The robot employs its arm‑mounted RGB‑D sensor to map a densely cluttered shelf. In the belief map~(bottom), white denotes high uncertainty, while solid colors show confidently mapped areas.
The robot first uses a next‑best‑view strategy to target uncertain regions on the right of the shelf.
Once those frontier objects are mapped, uncertain zones remain behind the cereal and Cheezit boxes. 
Our informed push sampler then most informative uncertain point (red dot), selects the cereal box since it occludes the area behind it, and pushes it (green arrow) to reveal the occluded space.}
\label{fig:cover_fig}
\vspace{-.5em}
\end{figure}

We build upon~\cite{marques25rss} and propose informed strategies for action sampling and selection in the MEM-task.
In particular, we propose a Reinforcement Learning (RL) based continuous NBV planner that uses predicted occupancy and semantic uncertainty to maximize expected information gain, enabling targeted exploration. 
RL enables the NBV planner to adapt to diverse, dynamic scenes and balances exploration–exploitation without crafted heuristics or high-dimensional POMDPs. 

Furthermore, we introduce uncertainty-informed push selection that targets occluding objects predicted to reduce uncertainty, choosing minimally invasive, high-impact manipulations to improve visibility of hidden regions.
Fig.~\ref{fig:cover_fig}~illustrates the two stages of our planning framework: First, the execution of a next‑best‑view followed by an uncertainty‑informed push to expose a highly uncertain region behind two view-blocking boxes.

Our experiments demonstrate that our novel uncertainty-informed MEM planner accurately maps cluttered scenes and outperforms the prior approach~\cite{marques25rss} in terms of action efficiency, i.e., it reduces the number of required push actions, and lowers the total number of object displacement, while being significantly faster in planning time.
An implementation of our method can be found on
Github\footnote{\url{https://github.com/HumanoidsBonn/manipulation_enhanced_map_prediction}}.

%%%%%%%%%%%%%%%%%%%%%%%%%%%%%%%%%%%%%%%%%%%%%%%%%%%%%%%%%%%%%%%%%%%%%%%%%%%%%%%%
\section{Related Work}
\label{sec:related}
\subsubsection{Interactive Perception}
Recent interactive perception frameworks that integrate physical interactions into the perception loop have demonstrated significant improvements in robustness and accuracy over active perception alone.
%By going beyond sensor repositioning to include intentional object manipulations and occlusion control, interactive perception systems can resolve ambiguities that passive and pure active approaches struggle with, yielding more generalizable and reliable scene understanding~\cite{yu2024manip} 
For example, Murali~\etal~report huge increase in object‐pose estimation accuracy compared to a baseline in dense clutter~\cite{murali2022active}. 
Similarly, Yao~\etal~demonstrated that manipulating leaves to reveal fruit markedly improves shape and pose estimates~\cite{yao2025icra}.  
Furthermore, interactive perception enables the autonomous, online construction of affordance‐based scene representations without requiring predefined object models~\cite{engelbracht2024spotlight}. 
However, these methods do not consider the problem of dense clutter at all,or in confined spaces, and approaches to uncover them. 

\subsubsection{Uncertainty-Aware Action Selection}
Uncertainty-aware action selection methods employ probabilistic models to inform perception and manipulation, yielding more efficient and reliable task execution than heuristic-driven strategies. 
For example, ActNeRF~\cite{dasgupta2024actnerf} quantifies 3D reconstruction uncertainty with NeRF ensembles to choose the next-best view or reorientation, accelerating modeling under occlusions versus random/greedy baselines.
Furthermore, for mobile manipulation, uncertainty can arbitrate between sensing and acting to reduce environmental and sensory uncertainty~\cite{tzes2022reactive}. Representing object pose as a distribution also enables assembly planners to sequence pushes, placements, and grasps to eliminate ambiguity, attaining higher success than deterministic methods~\cite{von2022uncertainty}.
Additionally, evidential neural networks~\cite{sensoy2018evidential} provide single-pass estimates of aleatoric and epistemic uncertainty for action selection. vMF-Contact~\cite{shi2024vmf} applies evidential vMF models for grasping in clutter, while Durasov~\etal~\cite{durasov2024uncertainty} use Dirichlet-based losses on 2D bird's eye view representations for 3D detection to identify out of distribution objects and improve detection.

\subsubsection{Manipulation‐Enhanced Mapping}
Robotic mapping in dense clutter requires strategic perception (\emph{where} to observe) and proactive interaction ({\emph{how} to manipulate occluding objects). Classical mechanical-search approaches exploit semantic relations to guide push-and-pick actions for target retrieval~\cite{sharma2023semantic,huang2022mechanical,danielczuk2019mechanical}. Interactive mapping extends this idea with manipulation primitives to resolve occlusions.
\mbox{Wang~\etal~\cite{wang2025curiousbot}} use an actionable 3D relational object graph for targeted interactions in mobile settings, whereas \mbox{Dengler~\etal~\cite{dengler23iros}} alternate next-best-view planning with push sampling once marginal map-entropy reduction saturates, for mapping of cluttered shelves.
More recently, Marques~\etal~\cite{marques25rss} adapted this idea and defined a POMDP solver whose belief resides in a metric‐semantic grid map, with evidential calibrated neural‐accelerated belief updates~(CNABU) to reason about object shapes and manipulation consequences, leading to higher map accuracy.

Building on \cite{marques25rss}, we retain its evidential belief predictors and explicitly exploit their occupancy and semantic uncertainty to learn continuous-space NBV planning and occlusion-critical push sampling. This extends MEM from fixed viewpoints and random, or scoring-based pushes to uncertainty-guided actions that focus on high information-gain regions, reduce redundant manipulation, and lower planning cost, thus enabling online semantic mapping in cluttered, dynamic scenes.

%Our work builds upon these lines of research by integrating uncertainty-aware action selection with learned continuous viewpoint planning and informed manipulation sampling rather than using random samples or push scoring networks. 
%Leveraging evidential networks \cite{sensoy2018evidential}, we explicitly model uncertainty in both occupancy and semantics.
%The uncertainty estimates generated by the evidential occupancy and semantic belief networks \cite{sensoy2018evidential}  guide a continuous RL‐based viewpoint planner and our novel occlusion‐critical push‐sampling module, which proposes safe, high‐value interactions. Compared to prior MEM~\cite{marques25rss} systems, our method focuses exploration on regions with high uncertainty reduction, reduces redundant manipulations, and lowers planning overhead—enabling real‐time semantic mapping in cluttered, dynamic scenes.

%%%%%%%%%%%%%%%%%%%%%%%%%%%%%%%%%%%%%%%%%%%%%%%%%%%%%%%%%%%%%%%%%%%%%%%%%%%%%%%%
\begin{figure*}[t]
\centering
\includegraphics[width=0.85\textwidth,trim={0cm 0cm 0cm 0cm}, clip]{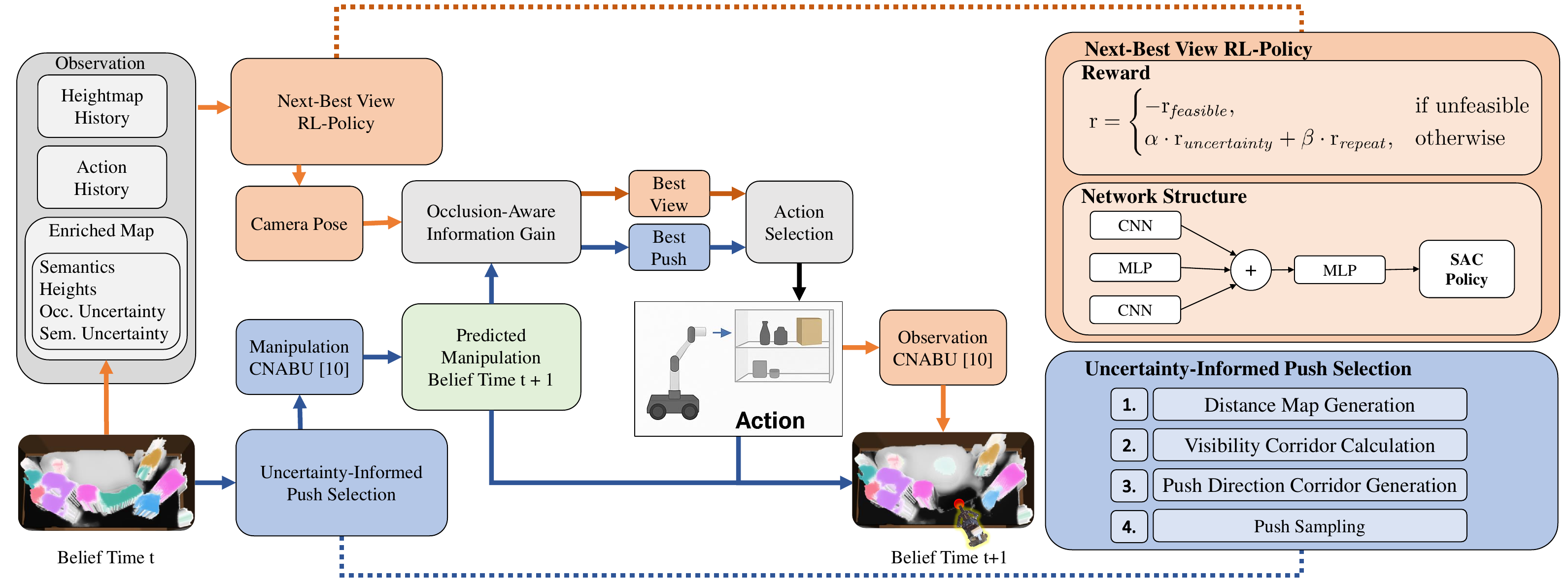}
%needed map t-1 side by side with real environment
% maybe use fov polygon in map to show the problem
\caption{Overview of our uncertainty-aware manipulation-enhanced mapping framework.
Our system integrates evidential semantic mapping with a reinforcement learning (RL) policy for continuous viewpoint planning (orange) and a strategic, uncertainty-informed push module~(blue). 
The RL agent operates in a 6D action space, leveraging uncertainty estimates from Beta and Dirichlet distributions to guide exploration towards occluded and ambiguous regions. 
When passive observation yields insufficient information gain, an informed manipulation policy selects occlusion-critical pushes to reveal hidden areas.}
\label{fig:vpp_overview}
\end{figure*}
\section{Our Approach}
%Efficient and accurate mapping in cluttered, occlusion-rich environments requires a unified strategy that combines perceptual inference with targeted interactions. 
%We address this challenge by integrating evidential metric‑semantic mapping~\cite{marques25rss} with continuous and adaptive viewpoint planning and uncertainty‑guided manipulation. 
%By explicitly modeling uncertainty in both occupancy and semantic predictions, our system dynamically selects camera poses and push actions that maximize information gain and minimize unwanted interactions. 
%This approach overcomes the limitations of fixed viewpoints, random manipulations, and heuristic planning, enabling faster, safer, and more complete scene reconstruction.
%The following subsections describe each component of our approach in detail.

%\label{sec:main}
%\subsection{Overview}
In this work, we address manipulation‐enhanced mapping~(MEM) of cluttered shelf environments using a robotic arm with a wrist‐mounted RGB-D sensor. 
An illustration of our pipeline is shown in Fig.~\ref{fig:vpp_overview}. 
Building on evidential metric-semantic mapping of Marques~\etal~\cite{marques25rss} via CNABU networks, we fuse RGB-D into occupancy (Beta) and semantic (Dirichlet) belief maps that encode predictions and their uncertainty. 
A \textbf{next-best view RL-policy} (orange) then selects $\mathrm{SE}(3)$ camera poses to observe regions of high-uncertainty. 
As active sensing shows diminishing returns, once the expected volumetric information gain drops below a threshold, an \textbf{uncertainty-informed push selector} (blue) identifies occlusion-critical objects using distance and visibility transforms, samples minimally invasive pushes, and estimates their information gain. 
An \textbf{action selector} (grey) then chooses sensing or manipulation by comparing expected gains, iterating until the mapping objective is met.
The individual parts of our system are now introduced in detail.

\subsection{Map Belief Representation}
\label{sec:uncertainty}
Accurate decision‐making in MEM‐tasks relies on reasoning about both observations and confidence. 
Following the CNABU framework~\cite{marques25rss}, we represent the map belief as a probabilistic estimate over occupancy and semantic categories, enriched with explicit uncertainty.
CNABU networks are designed to propagate belief in metric-semantic maps by predicting both semantic and occupancy beliefs, along with well-calibrated uncertainty estimates. 
They use Beta and Dirichlet distributions as conjugate priors for Bernoulli~(occupancy) and multinomial (semantic segmentation) likelihoods, respectively, ensuring that the resulting posteriors remain within the same distribution family. 
This property allows for closed-form estimation of expected values and predictive uncertainty, avoiding the computational overhead of sampling-based methods such as Monte Carlo dropout or deep ensembles.
Furthermore, they incorporate prior knowledge about object geometry, occlusions, and manipulation dynamics, enabling reliable forecasting of scene changes due to both observations and physical interactions.

The occupancy belief $O$ is a 3D voxel map of size \(H\times W\times D\), where each voxel is parameterized by \(\bm{\lambda}^O=(\alpha,\beta)\) and modeled by a Beta distribution whose predictive mean \(\mathbb{E}[\mathrm{Beta}(\alpha,\beta)]\) is the expected occupancy probability and whose variance 
$u_o = \mathrm{Var}[\mathrm{Beta}(\alpha,\beta)]$
quantifies epistemic uncertainty. 
The semantic belief $S$ is a 2D top‐down projected map of size \(H\times W\) with evidential parameters \(\bm{\lambda}^S\in\mathds{R}^{H\times W\times N_{\mathrm{classes}}}\), where \(\bm{\lambda}^S=(\lambda_1,\dots,\lambda_N)\) defines a Dirichlet distribution whose predictive class probabilities 
\vspace{-.5em}
\begin{equation}
\mathbb{E}[\mathrm{Dir}(\lambda_1,\dots,\lambda_N)]_n = \frac{\lambda_n}{S},\quad S=\sum_{n=1}^N\lambda_n,
\end{equation}
\vspace{-.5em}

\noindent yield a hard label \(\arg\max_n(\,\frac{\lambda_n}{S} )\), and whose uncertainty $u_s = N/S $
highlights low‐confidence regions~\cite{sensoy2018evidential}. 
In our framework, both \(u_o\) and \(u_s\) guide action selection to maximize expected information gain.
%Regions with high predictive uncertainty are prioritized for further exploration or targeted manipulation, enabling the agent to focus on parts of the scene that are both unknown and informative. 

\subsection{Next-Best View Selection}
To achieve efficient scene coverage, we couple CNABU’s one-step predictive mapping~\cite{marques25rss} with a reinforcement learning (RL) agent that operates directly on the evidential map representations. 
The agent receives occupancy and semantic beliefs together with their uncertainties and selects \(\mathrm{SE}(3)\) viewpoints to maximize expected information gain, guiding it toward the most informative, occluded regions. 

\subsubsection{\textbf{Action}}
We define the action space as a 6D vector \mbox{$(x_{cam}, y_{cam}, z_{cam}, x_{target}, y_{target}, z_{target})$}, with each component normalized to \mbox{$[-1, 1]$}.
At execution time these values are rescaled to predefined physical limits in SE(3).
In this action space, $cam$ represents the desired 3D position of the camera.
Furthermore, we represent orientation implicitly via a look-at $target$ to ensure the camera faces the region of interest (RoI), i.e.,  the shelf board.
%This simplifies learning by removing the need to learn the RoI from scratch.
%We define the action space as a 6D vector \mbox{$(x_{cam}, y_{cam}, z_{cam}, target_x, target_y, target_z)$}, where each component is normalized to the range \mbox{$[-1, 1]$}. 
%At execution time, these normalized values are rescaled according to predefined physical limits in the SE(3) space. 
%Rather than learning orientation directly through yaw, pitch, and roll angles, we represent orientation implicitly via a look-at target to ensure that the camera faces the region of interest (RoI) within the shelf, simplifying the learning problem by eliminating the need to learn the RoI from scratch.
\subsubsection{\textbf{Observation}}
Our observation space is detailed in Tab.~\ref{tab:observation_space} and it supports two key objectives:\\
\textbf{(i)~Enhanced world representation:} 
As described in Sec.~\ref{sec:uncertainty}, the agent's world representation comprises of the 3D occupancy and 2D semantic belief maps of known size. 
We project the thresholded occupancy belief into a 2D maximum-height map to reduce memory and computation while preserving occlusion reasoning and visibility assessment.  
Additionally, we incorporate cell-wise occupancy and semantic uncertainties, i.e., $u_o$ and $u_s$, from the Beta and Dirichlet distributions to enable the agent to learn how its actions affect scene uncertainty.\\ 
\textbf{(ii)~Action history awareness:} 
To avoid redundant viewpoints and encourage coverage diversity, the observation space includes the last $N_{\mathrm{hist}}$ executed actions along with their corresponding 2D height map observations.
This temporal context helps the agent identify previously explored areas and prioritize novel or more informative viewpoints.

%This structured observation space allows the agent to make more informed decisions, accelerating learning and improving viewpoint selection. 
% In total our obersave
% Our \todo{complete observation space} is shown in Tab~\ref{tab:observation_space}.

\begin{table}[t]
\centering
\resizebox{\linewidth}{!}{%
\begin{tabular}{lll}
\toprule
\textbf{Component} & \textbf{Content} & \textbf{Shape} \\
\midrule
\textbf{Enriched Map} & Maximum Height & $82 \times 157 px$ \\
& Semantic & $82 \times 157 px$ \\
 & Occupancy uncertainty $u_o$ & $82 \times 157 px$ \\
 & Semantic uncertainty $u_s$ & $82 \times 157 px$ \\

\textbf{Height Map History} & Past normalized height maps & $N_{\mathrm{hist}} \times 82 \times 157 px$ \\
\textbf{Action History} & Past normalized actions & $N_{\mathrm{hist}} \times 6$ \\
\bottomrule
\end{tabular}
}
\caption{Observation space used by the RL agent.}
\label{tab:observation_space}
\vspace{-5px}
\end{table}

\subsubsection{\textbf{Reward}}
The reward function comprises three terms with $w_1$ and $w_2$ as scaling factors:\vspace{-1em}%, divided in sparse and continuous reward signals, to facilitate stable and efficient training.
% \begin{equation}
% r = 
% \underbrace{r_{\text{feasibility}}}_{\text{sparse reward}}
% +
% \underbrace{\alpha \cdot r_{\text{uncertainty}} + \beta \cdot r_{\text{repeat}}}_{\text{continous rewards}}
% \end{equation}
\begin{equation}
r =
\begin{cases}
-r_{\text{feasibility}}, & \text{if unfeasible} \\
w_1 \cdot r_{\text{uncertainty}} + w_2 \cdot r_{\text{repeat}}, & \text{otherwise}
\end{cases}
\vspace{-1em}
\end{equation}
Given the continuous nature of the action space, sampled actions may be infeasible for execution. 
Hence, during training, we use the Klamp't motion planning library~\cite{hauser13klampt} to check for action feasibility. 
For infeasible actions, the agent is penalized via a negative sparse reward signal $r_{\text{feasibility}}$ and an empty partial observation is returned in the observation space to reflect the failed attempt.

% Furthermore, even feasible actions may yield only minor improvements to the map. 
% To guide the agent away from such suboptimal decisions, we introduce the penalty $r_{\text{bad\_view}}$, which signals that the action was feasible but not particularly informative. 
% Specifically, an action is classified as a bad view if less than $4\%$ of the perceived height map is occupied.

As \textbf{continuous reward} signals, we use the change in occupancy and semantic uncertainty \mbox{$r_{\text{uncertainty}} = \Delta u_o + \Delta u_s$} between two time steps.
Additionally, we use a repetition penalty $r_{\text{repeat}}$, which penalizes viewpoints close in position or orientation to previously visited ones~\cite{menon2023nbv}. 
In particular, $r_{\text{repeat}}$ is computed as follows:
\begin{align}
r_{\text{repeat}} &= -\sum_{(p_{\text{past}}, e_{\text{past}}) \in \mathcal{H}} \left( 
\gamma_p \frac{\theta_p - p_{\text{dist}}}{\theta_p} 
+ \gamma_e \frac{\theta_e - e_{\text{dist}}}{\theta_e}
\right), \\
p_{\text{dist}} &= \| p_{\text{curr}} - p_{\text{past}} \|, \quad
e_{\text{dist}} = \| e_{\text{curr}} - e_{\text{past}} \|,
\end{align}
Here, $ p_{\text{dist}}$ and $e_{\text{dist}}$ denote the Euclidean distance between positions and the cosine distance between the rotation vectors, respectively, within $\mathcal{H}$.
The terms $\gamma_p$ and $\gamma_e$ are scaling factors, while $ \theta_p $ and $\theta_e$ are the corresponding distance thresholds. 
The repetition penalty $r_{\text{repeat}}$ is computed only for entries in $ \mathcal{H}$ where $p_{\text{dist}} \leq \theta_p$ and~\mbox{$e_{\text{dist}} \leq \theta_e$}.

\subsection{Push Sampling}
\label{sec:push_sampling}
To enhance visibility and maximize information gain in cluttered and confined environments, particularly when only active sensing proves insufficient, we leverage strategic uncertainty-aware object manipulation to uncover occluded regions. 
In the following, we detail the process of identifying suitable manipulation targets, estimating visibility corridors, determining effective push directions, and finally sampling feasible push interactions. 
An illustration of each step is shown in Fig.~\ref{fig:push_pipeline}.

\subsubsection{\textbf{Target Locations}}
Unlike existing object-search methods \cite{bejjani2021occlusion,huang2022mechanical}, we define target locations as regions of the map with high semantic uncertainty. 
%i.e., where the robot has its least understanding of the environment.
We prioritize semantic uncertainty over occupancy uncertainty because even when occupancy estimates are confident, class labels can remain ambiguous. 
Focusing on semantics drives the robot to discover new objects and improve its understanding of known ones.
To identify these regions, we compute a distance transform over a thresholded version of the semantic uncertainty, where cells with a semantic uncertainty below $0.1$ are considered sufficiently certain and set to 0.
This value is empirically chosen to be optimal in terms of isolating regions whose certainty cannot be substantially improved given inherent network imperfections or sensor noise.
The distance transform assigns each cell a score based on how far it is from the nearest confidently classified cell~(either free or occupied), with the score scaled by cells occupancy uncertainty value.
From the resulting distance map, the a maximum of five cells with the highest distance values are selected as candidate target locations, while ensuring a minimum pairwise distance to promote spatial diversity as highlighted in Fig.~\ref{fig:push_pipeline:b}.

\begin{figure}[t]
\centering
% Plots below the legend
\subfloat[Current Map Belief\label{fig:push_pipeline:a}]{\includegraphics[width=.24\textwidth,trim={4cm 9.7cm 17cm 3.25cm},clip]{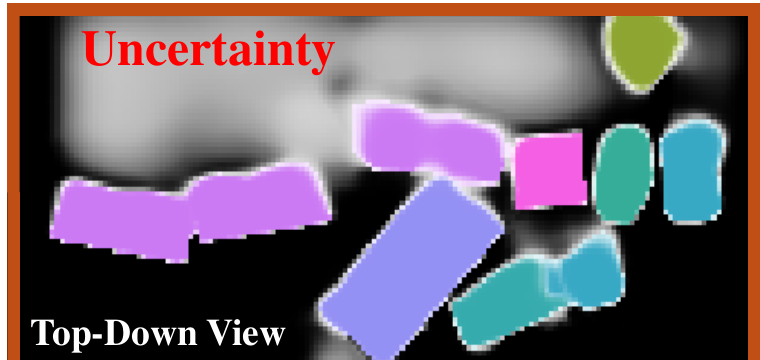}}
\hfill
\subfloat[Uncertainty-Based Distance Map\label{fig:push_pipeline:b}]{\includegraphics[width=.24\textwidth, trim={17cm 9.7cm 4cm 3.25cm},clip]{figures/Grafik_humanoids_push.pdf}}\\
 \vspace{-5px}
\subfloat[Visibility Corridor\label{fig:push_pipeline:c}]{\includegraphics[width=.24\textwidth, trim={4cm 3.25cm 17cm 9.5cm},clip]{figures/Grafik_humanoids_push.pdf}}
\hfill
\subfloat[Pushing Corridor With Samples\label{fig:push_pipeline:d}]{\includegraphics[width=.24\textwidth,trim={17cm 3.25cm 4cm 9.5cm},clip]{figures/Grafik_humanoids_push.pdf}}  
\caption{Uncertainty-informed push sampling. 
From the belief map (a; shelf borders in brown), we build an uncertainty-weighted distance field and sample target points (b).
Raycasting from each target to the shelf front yields visibility corridors and associated occluding objects (c; yellow).
For each occluder, we generate \(n_p\) feasible push proposals aligned with the corridor toward the clearing direction (d).
Afterwards, the manipulation CNABU evaluates the expected information gain of the push proposals.
}
\label{fig:push_pipeline}
\vspace{-10px}
\end{figure}

\subsubsection{\textbf{Visibility Corridor}}
For each selected target location, a visibility corridor is constructed to identify potential occlusions. 
This is achieved via raycasting from the target point towards the front of the shelf, defined by a fixed $y$ value and a variable $x$ value in the range [0, maximum width]. 
Along each ray, the number of semantic object instances from the one-hot-encoded map encountered is recorded, as well as the average occupancy probability, weighted by the corresponding epistemic occupancy uncertainty.

Neighboring rays that exhibit similar spatial structures (in terms of the order of object instances and non-occupied space) are clustered into unified corridors up to a maximum radius of $n_c$ cells. 
Each corridor is represented as a structure of its start point, width, length, order of objects, and average occupancy probability. 
These corridors are scored according to four criteria:
\vspace{-.5em}
\begin{equation}
    score = k_1 \cdot w + k_2 \cdot l + k_3 \cdot \bar{p}_{\text{occ}} + k_4 \cdot n_{\text{occ\_obj}}
\end{equation}
\vspace{-1.5em}

where $w$ is the width of the corridor, $l$ the normalized length of the visibility corridor, $\bar{p}_{\text{occ}}$ the average occupancy probability scaled by the normalized amount of unknown cells, $n_{\text{occ\_obj}}
$ the number of occluding objects, and $k_{1,2,3,4}$ are weights empirically found.
The object closest to the front of the shelf within the corridor with the highest score is selected as the manipulation target (see Fig.~\ref{fig:push_pipeline:c}). 
Note that if no occluding object is found, the target is considered observable through pure future viewpoint planning.

\subsubsection{\textbf{Pushing Corridor}}
The objective is to find a direction that minimizes secondary or inadvertent pushing of adjacent objects and prefers already mapped free areas over highly uncertain ones.
Therefore, unlike the visibility corridor, sectors with lower average occupancy and greater clearance are preferred to ensure good reachability and low collision probabilities with other objects. 
To determine viable pushing directions for the selected object, again raycasting is employed. 
Rays are radially cast from the centroid of the object identified as pushing target toward a circular boundary with a maximum radius of $n_c$ cells.
We identify object entities and their centroids by clustering connected regions of similar semantic labels in the current map.
For each object, we cast rays from its centroid inside the shelf in multiple directions, grouping these rays into sectors with an maximum angle of $30^\circ$. 
Each sector is evaluated based on the average occupancy along its rays and spatial constraints such as nearby obstacles or shelf boundaries (see Fig.~\ref{fig:push_pipeline:d}, blue corridor).

% \todo{We obtain object entities and centroids by clustering the semantic labels of the current map representation.
% These rays are grouped into directional sectors based on differences in average occupancy and spatial constraints.
% The length of the sector is penalized if it indicates collision risks, e.g., pushing toward a wall (see Fig~\ref{})} \todo{refer to Fig 3d}

\subsubsection{\textbf{Sampling Push Candidates}}
To generate possible pushes to execute, a starting corridor is established on the side opposite the pushing corridor (see Fig.~\ref{fig:push_pipeline:d}).
This is achieved by extending the lines from the end points of the corridor through the crossed contour intersection points, towards the opposite side of the contour.

Within this starting region, $n_p$ candidate push points are uniformly sampled. 
The associated push direction is computed as a linear combination of the pushing corridor vectors, weighted according to the relative position of each sampled point. 
Generating the push proposals in this fashion, results in feasible minimally invasive push actions aligned with free-space directions.

\subsection{Manipulation-Enhanced Planning}
We combine our viewpoint selection policy with the uncertainty informed push-based manipulation strategy to realize a complete manipulation-enhanced mapping pipeline.
While the RL-based next-best view planning agent seeks to maximize information gain through passive camera movement, physical occlusions can make certain regions permanently inaccessible to observation.
In such cases, the system queries the push sampling module to actively remove occlusions.

For decision making, we employ a variant of the occlusion-aware Volumetric Information Gain~(VIG) method proposed by~\cite{marques25rss}. 
For the NBV module, we query the trained RL agent to generate a single candidate action, obtain the predicted camera pose, and compute the associated $\text{VIG}_{nbv}$ based on the evidential occupancy prediction at time $t-1$.
Similarly, for the push sampling module, we simulate the outcome of each push by generating its parameterization as described in \cite{marques25rss} and input it together with the evidential occupancy prediction at time $t-1$ to the manipulation-CNABU to predict the resulting updated evidential map.
Subsequently, we query the NBV agent on the post-push map, incorporating the effects of the simulated manipulation, and compute the corresponding $\text{VIG}_{push}$.

The system then compares the maximum expected $\text{VIG}_{nbv}$ from viewpoint planning against $\text{VIG}_{push}$.
If $\text{VIG}_{nbv}$ exceeds that of any proposed push by a predefined margin \mbox{$\text{VIG}_{nbv} \cdot \Delta_\text{view} > \text{VIG}_{push}$}, the system executes the corresponding NBV action.
Otherwise, if pushing is expected to yield higher information gain, the robot proceeds to execute the selected push to actively modify the scene and takes its predicted belief generated by the manipulation-CNABU as map representation. This strategy ensures that executing viewpoints is always prioritized when sufficient information gain can be achieved without manipulation, while empowering the system to physically interact with the environment whenever necessary to uncover previously inaccessible regions and maximize overall map accuracy.

\begin{table}[t]
\centering
\begin{tabular}{ll|ll}
\hline
\textbf{Parameter} & \textbf{Value} & \textbf{Parameter} & \textbf{Value} \\
\hline
$N_{\mathrm{hist}}$ & 4 & $\Delta_{\text{view}}$ & 2 \\
$\theta_{\text{sem}}$ & 0.65 & $\theta_{\text{occ}}$ & 0.87 \\
$\gamma_p$ & 0.5 & $\gamma_e$ & 0.5 \\
$\theta_p$ & 0.1\,m & $\theta_r$ & 0.034\,rad \\
$w_1$ & 10 & $w_2$ & 2 \\
\hline
\multicolumn{4}{c}{\textit{Pushing Parameters}} \\
\hline
$k_1$ & 2 & $k_2$ & 3 \\
$k_3$ & 4 & $k_4$ & 5 \\
\hline
\end{tabular}
\caption{Hyperparameters for RL training and push sampling.}
\label{tab:compact_hyperparams}
\vspace{-1em}
\end{table}
\section{Experimental Evaluation}
\label{sec:exp}
We perform three experimental evaluations to showcase the performance of our next-best view and push selection approach. 
First, we present quantitative simulation results that highlight our pipeline’s improvements in mapping accuracy, compared to the state-of-the-art for manipulation-enhanced mapping~\cite{marques25rss}.
Next, we present several ablations of our method to highlight the influence of uncertainty measurements for action selection and evaluate different versions of our novel push sampling strategy.
Finally, we show a qualitative experiment in a real hardware setup in zero-shot fashion in the supplementary video\footnote{\url{https://youtu.be/9h2MKLzQl80}}.
%Finally, we study the robustness of our system in terms of its zero-shot transferability to a physical setup.
\begin{figure*}[t]
\centering

% Legend at the top
\captionsetup[subfigure]{labelformat=empty}
\subfloat[\label{fig:sim_eval_non_pushing_label}]{\includegraphics[width=\textwidth,trim={0cm 2.2cm 0cm 2.75cm},clip]{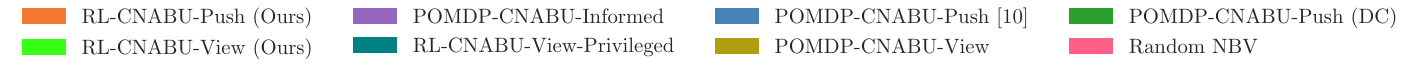}}  \\
\vspace{-20px}
% Plots below the legend
\subfloat[(a) Mapping - Occupancy\label{fig:sim_eval:b}]{\includegraphics[width=.24\textwidth, trim={0.2cm 0cm 0.8cm 0.2cm},clip]{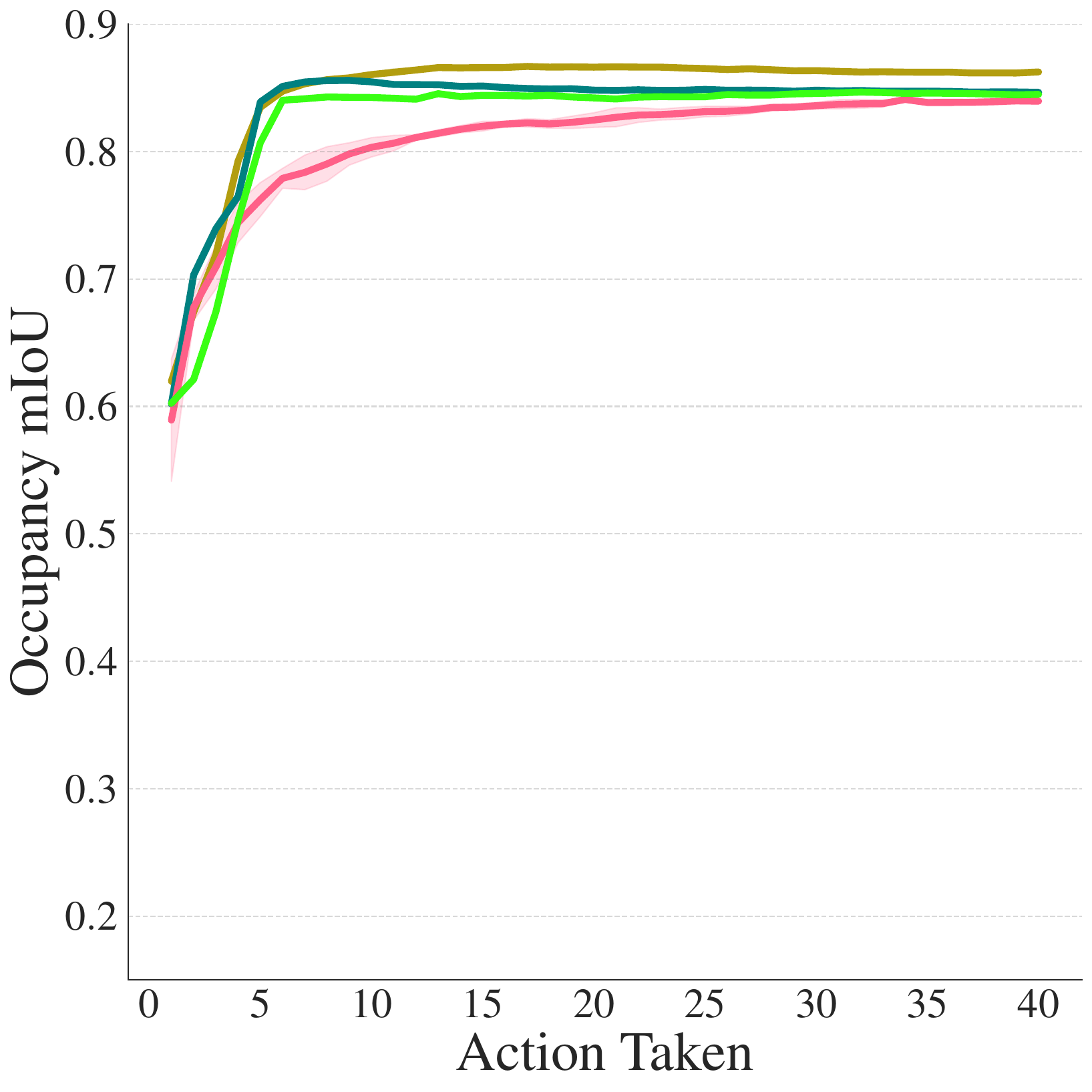}}
\hfill
\subfloat[(b) MEM - Occupancy\label{fig:sim_eval:a}]{\includegraphics[width=.24\textwidth,trim={0.2cm 0cm 0.8cm 0.2cm},clip]{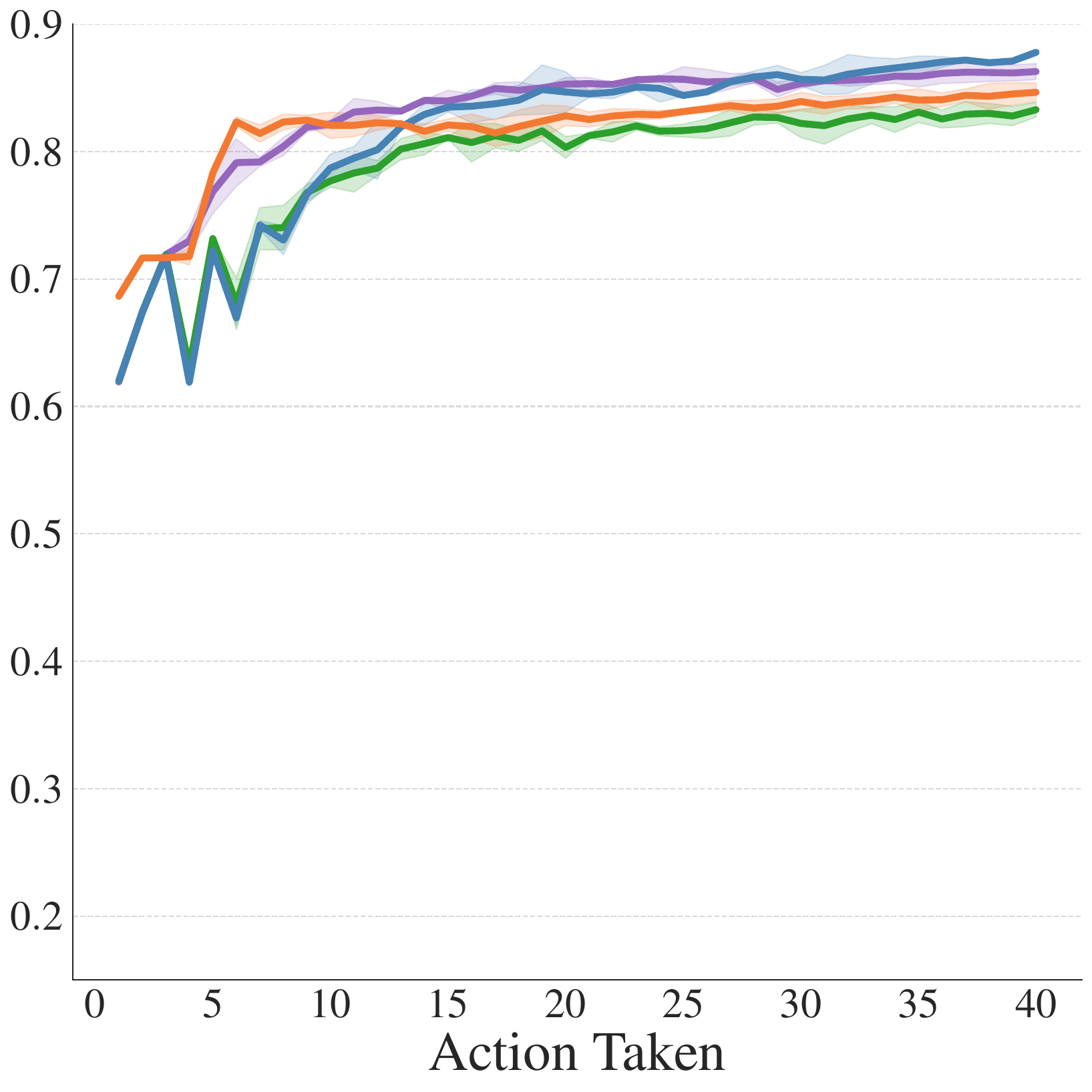}}
\hfill
\subfloat[(c) Mapping - Semantics\label{fig:sim_eval:c}]{\includegraphics[width=.24\textwidth,trim={0.2cm 0cm 0.8cm 0.2cm},clip]{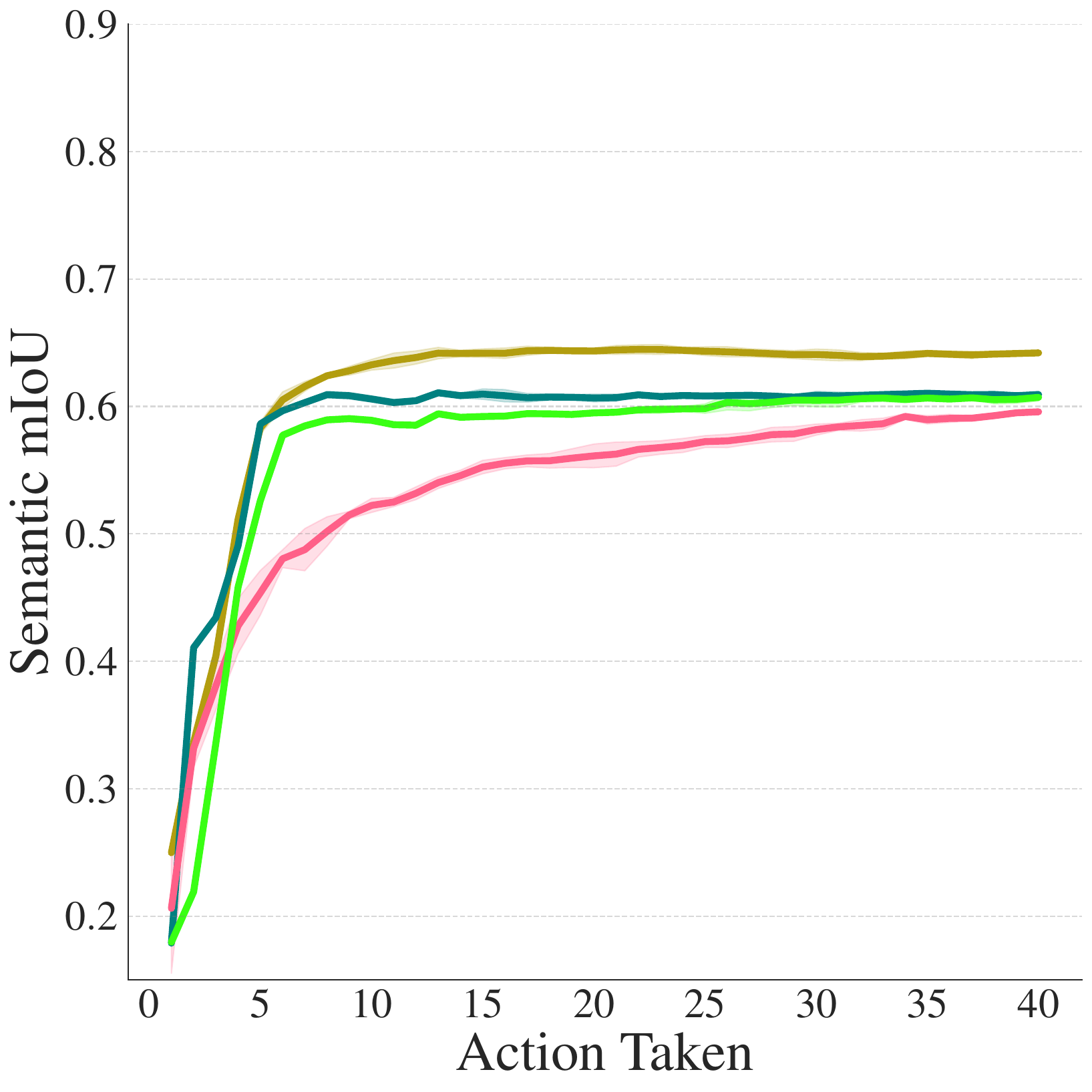}}  
\hfill
\subfloat[(d) MEM  - Semantics\label{fig:sim_eval:d}]{\includegraphics[width=.24\textwidth, trim={0.2cm 0cm 0.8cm 0.2cm},clip]{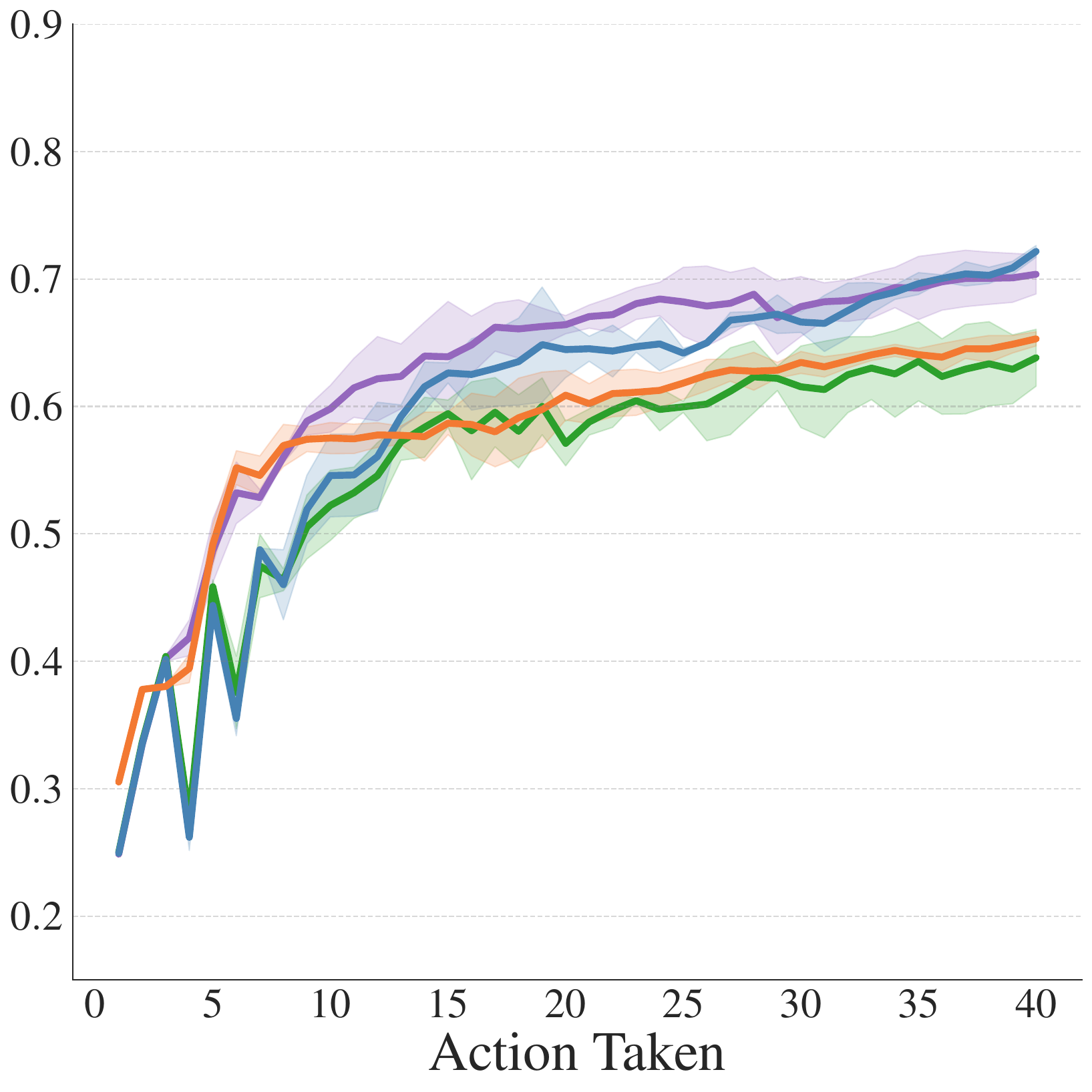}}
\caption{
Simulation results across 25 highly cluttered shelf environments with an action budget of 40 steps for RL-CNABU-Push and RL-CNABU-View against SOTA in active and interactive perception methods, showing both occupancy and semantic IoUs over time for each method. Standard deviation is represented as shading around each plot. 
As can be seen, our uncertainty-informed push strategy, POMDP-CNABU-Informed, outperforms previous work in terms of stable mapping without object drops. Furthermore, our proposed RL-CNABU-View is capable of achieving sufficient mapping behaviour without relying on privileged knowledge.
}
\label{fig:simulation eval}
\end{figure*}

\subsection{Experimental Setup and Training}
For simulation, we use the PyBullet physics engine~\cite{coumans2021pybullet} with a 6-DOF UR5 robotic arm, equipped with a \mbox{Robotiq 2F-85} gripper and a Realsense L515 RGB-D camera mounted on its end-effector. 
The simulation matches the real-world camera parameters and all experiments are trained and evaluated on an NVIDIA RTX 4080 Super 16GB GPU.
To train the continuous viewpoint planning agent, we use the Soft Actor-Critic (SAC) algorithm~\cite{haarnoja2018soft}, implemented via the Stable Baselines3 framework~\cite{stable-baselines3}. 
%Our custom environment extends the MEM-simulation from Marques~\etal~\cite{marques25rss} for our RL pipeline.

For the actor and critic networks, we use the following custom feature extractor: (i) A convolutional encoder for the enriched map, consisting of three convolutional layers with 16, 32, and 64 filters, each followed by LeakyReLU activation, with the last layer performing global average pooling and flattening. 
(ii) A similar encoder for the stacked height maps.
(iii) A fully connected module embedding the last four actions into a vector of size 64.
The outputs are concatenated and passed through an MLP, projecting it to a 128-dimensional latent space, feeding both policy and critic heads. 
Training uses a batch size of 256, a replay buffer of $10^5$ transitions, and a learning rate of $3\cdot10^{-4}$.

The agent is trained for $500,000$ iterations across approximately $12,500$ procedurally generated shelf configurations. 
Object counts range from 15 to 30 per scene. 
To simulate varying occlusion levels for training, large objects are biased toward the front of the shelf and smaller ones toward the back, increasing visual complexity. 
The action space is sampled from a Cartesian space of $\SI{0.8}{\meter} \times \SI{0.2}{\meter} \times \SI{0.2}{\meter}$ outside and inside the shelf, respectively. 
The hyperparameters used for all experiments are stated in Tab.~\ref{tab:compact_hyperparams}.

% $N_{\mathrm{hist}}=4$, $\Delta_\text{view} = 2$, $\theta_{\text{sem}} = 0.65$, and $\theta_{\text{occ}} = 0.87$.
% for the pushing we use $k1=2$, $k2=3$, $k3=4$, $k4=5$
% For the reward function of the viewpoint planning agent, we set $w_1 = 10$, $w_2 = 2$, $\gamma_p= \gamma_e= 0.5$, $\theta_p = 0.1\, m$, and  $\theta_p = 0.034\,rad $. %\todo{I think there are more parameters for the pushing}

% During training, we sample diverse shelf object configurations, including both high- and low-occlusion scenes. 
% High-occlusion configurations are generated by preferentially placing larger objects toward the front of the shelf, while smaller objects are more likely positioned toward the back, thereby increasing scene complexity and visual clutter.

% The Cartesian sampling space spans~$\SI{0.8}{\meter}\times\SI{0.2}{\meter}\times\SI{0.2}{\meter}$ outside and inside the shelf for the camera's position and a target point, respectively.
% The reinforcement learning agent is trained for 500,000 iterations across approximately 12,500 procedurally generated shelf configurations, with the number of placed objects randomized in each configuration between 15 and 30.

% The following hyperparameters are used for all experiments:
% $N_{\mathrm{hist}}=4$, $\Delta_\text{view} = 2$, $\theta_{\text{sem}} = 0.65$, and $\theta_{\text{occ}} = 0.87$.
% For the reward function of the viewpoint planning agent, we set $\alpha = 10$ and $\beta = 2$.

\subsection{Metrics and Baselines}
\label{sec:metrics}
To evaluate our informed action selection pipeline for the given MEM-task, we evaluate the following metrics for semantic and occupancy accuracy, object displacement, and pipeline efficacy:
\begin{itemize}
    \item \textbf{Timings:} Inference and execution timings of the pipelines 
    \item \textbf{mIOU:} Semantic and occupancy mean Intersection-over-Union, computed as the average per-class IoU between predictions and ground truth, calculated on the CNABU network output after transformation into the map representation.
    \item \textbf{mAD:} Mean displacement of add objects averaged over all evaluated scenes.
    \item \textbf{NumPush:} Average number of pushes per scene.
    \item \textbf{Collisions:} Average percentage of collisions occurred over all evaluated scenes.
\end{itemize}
Furthermore, we compare our  \textbf{full} system \textbf{RL-CNABU-Push (Ours)} against the following baselines and state-of-the-art:
\begin{itemize}
    \item \textbf{RL-CNABU-View (Ours):} Uncertainty-aware RL based next-best-view (NBV) planner. 
    \item \textbf{RL-CNABU-View-Privileged (Ours):} RL based NBV planner that uses ground truth occupancy and semantics from privileged simulation information for rewards.
    \item \textbf{Random-View:} View planner that randomly selects sensor poses from the RL action space. 
    \item \textbf{POMDP-CNABU-Push:} Full pipeline by Marques~\etal~\cite{marques25rss} that uses a partially observable Markov decision process (POMDP) two-step look-ahead solver with randomly sampled push actions.
     \item \textbf{POMDP-CNABU-Push (ToF):} POMDP-CNABU-Push terminated when any of the objects falls over. 
    \item \textbf{POMDP-CNABU-View:} Only NBV planning by~\cite{marques25rss}.
    \item \textbf{POMDP-CNABU-Informed:} Pipeline~\cite{marques25rss} with novel push sampling (see Sec.~\ref{sec:push_sampling}).
\end{itemize}
For each method, we perform three evaluation trials and report the mean and the standard deviation. 

\subsection{Quantitative Experiments}
 We present our quantitative experiments, designed to evaluate the effectiveness of our proposed manipulation-enhanced mapping pipeline. 
 To compare our approach against the POMDP-CNABU planners~\cite{marques25rss}, we re-implemented their pipeline and used the provided CNABU-models from their code release. 
 In order to use the models for our approach, no fine-tuning on the novel set of continuous viewpoints was performed. 
However, we increased the certainty threshold for full map completion after which no pushes are performed from 95\%, as stated in \cite{marques25rss}, to 99\%. 
This increases the chance of further pushes in the long run and shows the effect of successful push prediction on the over- or under-confidence of the CNABU-models.

To evaluate the effectiveness of our system in creating accurate scene representations, we assess the quality of semantic and occupancy maps across 25 highly cluttered shelf environments with an action budget of 40 steps. We evaluate them according to the metrics specified in \mbox{Sec.~\ref{sec:metrics}}.

\begin{table}[t!]
	\centering
	\resizebox{\linewidth}{!}{
\begin{tabular}{l|c|c|}
        \textbf{Metric} & \textbf{POMDP-CNABU-Push} & \textbf{RL-CNABU-Push} \\
        \hline
        \textbf{View Point Selection Time (s)} & $7.632 \pm 1.892$ & $\textbf{0.022}\pm 0.001$ \\
        \textbf{Push Selection Time (s)}       & $31.199 \pm 7.814$& $\textbf{0.976} \pm 0.618$ \\
        \textbf{Full Iteration Time (s)}       & $40.093 \pm 10.095$& $\textbf{1.656} \pm 1.117$ \\
\end{tabular}
}
    \caption{Timing performance in seconds across the 25~evaluation scenes averaged over 40-step rollouts and across three trials. Our RL and uncertainty aware push approach highly reduces in general the inference times of the task.}
	\label{tab:timing}
    \vspace{-10px}
\end{table} 

\subsubsection{\textbf{Timings}}
\label{sec:timings}
As shown in Table~\ref{tab:timing}, our method significantly reduces decision latency in both viewpoint and manipulation selection.
This is primarily due to our reinforcement learning-based viewpoint planner, which operates in a continuous action space and avoids exhaustive enumeration and ray-casting of discrete candidates, as well as our informed push sampling strategy, which narrows down the set of evaluated push actions while preserving their impact on map improvement.
In particular, our \mbox{\textbf{RL-CNABU-View}} planner achieves a \textbf{99.7}\% reduction in average inference time per step compared to the original \mbox{\textbf{POMDP-CNABU-View}} planner.
Similarly, the informed push selection time and the total per-iteration execution times are significantly reduced, enabling faster scene coverage and improved responsiveness.

\begin{table*}[ht!]
\centering
\resizebox{\linewidth}{!}{
\begin{tabular}{l|c|c|ccccc}
\textbf{Model} & \textbf{Perception} & \textbf{Push Strategy}  & \textbf{Occ. mIOU} $\uparrow$ & \textbf{Sem. mIOU} $\uparrow$ & \textbf{mAD [m]} $\downarrow$ & \textbf{NumPush [\#]} $\downarrow$ & \textbf{Collisions [$\%$]} $\downarrow$\\
\hline
RL-CNABU-View (Ours) & Active& None & $0.844 \pm 0.001$ & $0.607 \pm 0.002$ & - & - & -  \\
RL-CNABU-View-Privileged & Active& None & $0.846 \pm 0.001$ & $0.609 \pm  0.002$ & - & - & -  \\
POMDP-CNABU-View~\cite{marques25rss} & Active& None & $0.862 \pm 0.001$ & $0.641 \pm 0.001$ & - & - & - \\
Random-View & Active& None & $0.839 \pm 0.002 $ & $0.595 \pm 0.001$ & - & - & - \\
\hline
RL-CNABU-Push (Ours) & Interactive & $u_s|u_o$ & $0.846 \pm 0.007$ & $0.653 \pm 0.005$ & $\bf0.995 \pm 0.049$ & $ \bf{2.59 \pm 1.24}$ &  \textbf{6.67\%} \\
POMDP-CNABU-Push~\cite{marques25rss} & Interactive& Random Sampling  & $\bf{0.877 \pm 0.003}$ & $\bf{0.720 \pm 0.019}$ & $1.927 \pm 0.046$ & $5.25 \pm 2.61$ & 25.33\% \\
POMDP-CNABU-Push (ToF) & Interactive& Random Sampling & $0.832 \pm 0.005$&$0.638 \pm 0.022$&$1.841 \pm 0.033$&$4.65 \pm 2.50$&25.33\%\\

POMDP-CNABU-Informed & Interactive & $u_s|u_o$ & $0.862 \pm 0.006$ & $0.703 \pm 0.015$ & $1.056 \pm 0.157$ & $2.61 \pm 1.49$ & \textbf{6.67\%} \\
% POMDP-CNABU-Informed & Ablation & $u_s|u_{bin}$  & $0.862 \pm  0.015$ & $0.695 \pm 0.018$ & $0.828  \pm 0.109$ & $2.17 \pm 1.61$ & \textbf{5.33\%}\\
POMDP-CNABU-Informed & Interactive & $u_o|u_o$ & $0.862  \pm 0.005$ & $0.698  \pm 0.012$ & $1.060  \pm 0.255$ & $2.91 \pm 1.24$ & 9.33\% \\
\end{tabular}
}
\caption{Quantitative comparison over 25 cluttered scenes, averaged across three trials.
We report occupancy and semantic mIoU, mean action displacement (mAD), push count, and collision rate.
Our informed strategies ($u_s|u_o$, $u_s|u_b$, $u_o|u_o$) and RL-based methods show benefits of uncertainty-aware action selection. 
While POMDP-CNABU-Push achieves the highest semantic mIoU, it does so at the cost of more frequent pushing and a high collision rate.
In contrast, our RL-CNABU-Push approach offers a better balance, achieving competitive mapping performance with fewer pushes, lower object displacement, and reduced number of collisions.}
\label{tab:cquantitative}
\vspace{-5px}
\end{table*}

\subsubsection{\textbf{Map Quality}}
\label{sec:map_qual}
The results in terms of map quality across the whole action budget of 40 steps is illustrated in Fig.~\ref{fig:simulation eval} with a more detailed analysis of the last step shown in Tab.~\ref{tab:cquantitative}. In the following, we summarize the main results.

Among active perception methods, although \mbox{\textbf{POMDP-CNABU-View}} performs better than \mbox{\textbf{RL-CNABU-View}} in mapping accuracy, it highly relies on the quality of the predefined fixed views and is computationally expensive as it uses a exhaustive greedy search over up to 300 fine-tuned views, querying expected information gain at each step.  
In contrast, our \mbox{RL-CNABU-View}  predicts the next-best view with a single forward pass through a learned continuous-space policy, substantially reducing computation time, as discussed in Sec.~\ref{sec:timings}, while maintaining comparable mIoU.

With regards to active versus interactive perception, Tab.~\ref{tab:cquantitative} shows that the worst performing interactive perception method achieves higher semantic mIOU than the best active perception method, highlighting the need for physical interaction. 
In particular, our uncertainty-informed push method, \mbox{\textbf{RL-CNABU-Push}} achieves higher semantic mIoU than both \mbox{RL-CNABU-View} and \mbox{POMDP-CNABU-View}, demonstrating that targeted push actions can improve mapping accuracy by uncovering hidden regions.
For interactive perception, \figref{fig:simulation eval} and \tabref{tab:cquantitative} show that \mbox{\textbf{POMDP-CNABU-Push}} achieves higher mIOU metrics compared to \mbox{RL-CNABU-Push}. 
However, unlike all other methods, this proposed pipeline~\cite{marques25rss} does not consider any early stopping when an object falls over, e.g., out of the shelf, due to collisions. 
When POMDP-CNABU-Push is terminated on an object fall, referred to as POMDP-CNABU-Push~(ToF), our proposed \mbox{\mbox{\textbf{RL-CNABU-Push}}} achieves higher occupancy and semantic mIOU compared to POMDP-CNABU-Push~(ToF).

As all \mbox{POMDP-CNABU-Push} variants use random push sampling to generate pushes without knowledge of effective pushing corridors, they result in the highest mean displacement and collision rates among interactive perception methods.  
In \mbox{\textbf{POMDP-CNABU-Informed}}, replacing random sampling with our uncertainty-informed strategy increases mIoU while significantly reducing average displacement, number of pushes, and collisions.  
The reduced displacement and collisions in all the informed push methods confirm that our interactions are minimally invasive compared to the random push sampling baseline. 
Furthermore, informed push sampling methods require only 2–3 pushes versus 4–5 in random methods to achieve comparable or better map quality, demonstrating more targeted uncovering of hidden spaces.  
%Thus, our uncertainty-informed approach reduces critical failures by up to \textbf{78.96\%}, resulting in more stable interactions across cluttered scenes.
%Although the original \mbox{\textbf{POMDP-CNABU-Push}} pipeline~\cite{marques25rss} is modular, replacing its random push and discrete-view strategies with our methods improves overall performance.  
%Finally, our full system (\mbox{\textbf{RL-CNABU-Push}}) outperforms all baselines in both semantic and occupancy mIoU, indicating more complete and accurate reconstructions. 
%Furthermore, it achieves this with fewer average pushes compared to \mbox{\textbf{POMDP-CNABU-Informed}}, demonstrating the advantage of our uncertainty-guided push selection strategy. 
These results confirm that leveraging uncertainty for both viewpoint and manipulation planning leads to comparable or better quality map quality, particularly in scenes with high occlusion.
Furthermore, as the timing results indicate, our method achieves comparable mapping accuracy, with significantly reduced computational cost.

% reduces the time needed by $\textbf{96.87\%}$. The total per-iteration execution time, including both planning and simulation overhead, is reduced by approximately $\textbf{95.87\%}$, enabling faster scene coverage and improved responsiveness.
\subsection{Influence of Uncertainty}
To assess the influence of uncertainty metrics on the action selection process, we perform two ablation studies. 
\subsubsection{\textbf{Next-Best View planning}}
We evaluate our uncertainty-aware agent  \mbox{\textbf{RL-CNABU-View}} against \mbox{\textbf{RL-CNABU-View-Privileged}}, which incorporates ground-truth occupancy and semantics into its reward, using the same test set as in Sec.~\ref{sec:map_qual}.  
We adopt the reward \(r = r_{\text{IoU}} + r_{\text{repeat}}\), where \(r_{\text{IoU}}\) is the change in semantic and occupancy IoU from the ground-truth map.  
This ablation shows that uncertainty-based rewards suffice to learn effective next-best view selection without privileged information.  
We also compare our \mbox{RL-CNABU-View} method to \mbox{\textbf{Random-View}} to verify that our RL agent learns informative views. 

As shown in Fig.~\ref{fig:simulation eval}.a,c, our reward achieves similar or only slightly reduced long-term performance compared to the privileged agent.  
These findings demonstrate that uncertainty is a reliable performance proxy when ground-truth data is unavailable.  
Moreover, \mbox{Random-View} shows lower occupancy and semantic mIOU gains in the initial phases in \figref{fig:simulation eval}, with the final mapping accuracy also the lowest among all active perception methods. 
Overall, our RL agent attains higher IoU, confirming its learned ability to select informative views.  

\subsubsection{\textbf{Manipulation}}
To evaluate the influence of different uncertainty representations on our push sampling strategy, we conduct ablation experiments with two variants for distance map and pushing corridor generation: %\todo{three variants}:
\textbf{(i)} using semantic uncertainty for distance map generation while using occupancy uncertainty for the pushing corridor ($\bf{u_s|u_o}$) to highlight semantically ambiguous but structurally visible regions, providing complementary cues for selecting informative push targets, the proposed method of this paper; and \textbf{(ii)} using occupancy uncertainty for both the uncertainty-based distance map and the pushing corridor generation ($\bf{u_o|u_o}$),
%;  and \textbf{(iii)} using a binary uncertainty map $u_{\text{bin}}$, derived from $u_o$, for pushing corridor generation ($\bf{u_o|u_{\textbf{bin}}}$), to isolate the effect of per-cell uncertainty magnitudes. In this binary variant, all uncertain occupancy cells above a threshold of $0.01$ are assigned a value of 1, and all others are set to 0.
We performed this ablation on the \mbox{\textbf{POMDP-CNABU-Informed}} push planner to have a more comparable evaluation that only focuses on the influence of the push selection, due to its systematic greedy search characteristics.% of the predefined fixed viewpoints.

The results after the final action step are presented in~Tab.~\ref{tab:cquantitative}. 
%While the ($u_o|u_{\text{bin}}$) variant achieves the lowest overall collision rate and average number of pushes, it also results in the lowest IoU values among the three variants. 
%Given the consistently low mAD, we hypothesize that this behavior stems from the method being overly cautious, i.e., it avoids pushing anywhere other than into clearly free space, which limits opportunities for impactful push candidates. Nonetheless, this variant may be suitable when minimally invasive interactions are preferred.
As seen from the last two rows, using  $u_s$  as distance map yields marginally higher semantic mIOU as it is better able to target regions with high semantic uncertainty. 
Furthermore, it leads to lower collision rates possibly due to clustering of semantically similar cells leading to pushes being executed on single objects reducing secondary pushes. 

Our ablation studies on the influence on the uncertainty metrics show that uncertainty is a good approximation for action selection between active sensor placement and manipulation. While semantic uncertainty can be used to target regions for active observation as well as for the push distance map, occupancy uncertainty can be effectively used to generate minimally invasive push actions. 

%\subsection{Hardware Experiments}
%%%%%%%%%%%%%%%%%%%%%%%%%%%%%%%%%%%%%%%%%%%%%%%%%%%%%%%%%%%%%%%%%%%%%%%%%%%%%%%%
\section{Conclusion}
\label{sec:conclusion}

In this paper, we presented a novel framework for efficient manipulation-enhanced semantic mapping  that combines evidential metric-semantic map belief updates with uncertainty-aware action selection. 
By leveraging uncertainty estimates derived from the Beta and Dirichlet distributions for occupancy and semantics respectively, our system effectively selects informative next-best views through reinforcement learning and targets occlusion-critical objects for minimally invasive manipulations. 
As we showed in the experiments,  out framework enables a robot to construct complete semantic maps in cluttered, occlusion-heavy shelf environments.

We demonstrated that our uncertainty-informed push planning approach outperforms the state-of-the-art method in terms of object displacement, action efficiency, and reduced number of fallen objects. It also shows comparable behavior in terms of map accuracy.
Furthermore, our approach is highly computationally efficient due to the reinforcement learning based viewpoint planning and the informed push sampling strategy. 
These results highlight the importance of principled uncertainty-informed action selection for interactive perception in cluttered and confined spaces.
 
\section{Acknowledgments}
We would like to thank Joao Marcos Correia Marques and Murad Dawood for the fruitful discussions.

%%%%%%%%%%%%%%%%%%%%%%%%%%%%%%%%%%%%%%%%%%%%%%%%%%%%%%%%%%%%%%%%%%%%%%%%%%%%%%%%
% Future work only if it makes sense 
%\emph{Future work: Use only if applicable -- but if so, use the following sentence to start:} 
%
%Despite these encouraging results, there is further space for
%improvements. For example, ...

%%%%%%%%%%%%%%%%%%%%%%%%%%%%%%%%%%%%%%%%%%%%%%%%%%%%%%%%%%%%%%%%%%%%%%%%%%%%%%%%
% Only if applicable
%\section*{Acknowledgments}
%We thank XXX for fruitful discussions and for ...
\bibliographystyle{IEEEtran}
\bibliography{bibliography}

\end{document}